\documentclass{aircc}
\usepackage{mathpartir}
\usepackage{hyperref}
\usepackage{listings}
\usepackage{epsfig,fullpage,multirow,amsmath,amsfonts,latexsym,mathrsfs} 
\usepackage{times}
\usepackage{graphicx}
\usepackage{float}
\usepackage{bm}
\usepackage{nopageno}

\usepackage{mwe} 
\usepackage{wrapfig}

\begin{document}

\title{
A Deep Learning System for \\
Domain-specific Speech Recognition
}

\author{Yanan Jia}
\affiliation{Businessolver \\ \email{\url{yjia@businessolver.com}}}

\maketitle


\begin{abstract}   
As human-machine voice interfaces provide easy access to increasingly intelligent machines, many state-of-the-art automatic speech recognition (ASR) systems are proposed. However, commercial ASR systems usually have poor performance on domain-specific speech especially under low-resource settings. The author works with pre-trained DeepSpeech2 and Wav2Vec2 acoustic models to develop benefit-specific ASR systems. The domain-specific data are collected using proposed semi-supervised learning annotation with little human intervention. The best performance comes from a fine-tuned Wav2Vec2-Large-LV60 acoustic model with an external KenLM, which surpasses the Google and AWS ASR systems on  benefit-specific speech. The viability of using error prone ASR transcriptions as part of spoken language understanding (SLU) is also investigated. Results of a benefit-specific natural language understanding (NLU) task show that the domain-specific fine-tuned ASR system can outperform the commercial ASR systems even when its transcriptions have higher word error rate (WER), and the results between fine-tuned ASR and human transcriptions are similar.
\end{abstract}



\section{Introduction}
\label{se:intro}
Speech input industrial applications, such as smart voice assistants and customer service voice chat-bots, offer obvious benefits to users in terms of immediacy, touch-free interaction, and convenience via language learning.
Automatic speech recognition (ASR) systems have been deployed as an input method in many successful commercial products and have become a popular human-machine interaction modality.

Building ASR systems typically requires a large volume of training data to cover all possible factors contributing to the creation of speech signals, including but not limited to demographic variety, noise conditions, emotional state, topics under discussion, and the language used in communication. Data has been central to the success of end-to-end speech recognition. At higher-resource conditions, many state-of-the-art ASR pipelines approach or exceed the accuracy of human workers on several benchmarks. However, the ASR performance gap between different datasets remains \cite{Ragni2014DataAF}. Most acoustic models transfer poorly to domain-specific speech especially under low-resource settings \cite{hsu}.
 
 
DeepSpeech2 (DS2) and Wav2Vec2  are popular pre-trained acoustic models (AMs).  ASR performance on domain-specific speech can be significantly improved by fine-tuning pre-trained AMs on the in-domain data along with external in-domain language models (LMs). 
However, even for fine-tuning, to achieve improved accuracy, large volumes of high-quality annotated data are required.  

The data resources used for this experiment consist of an artificial dataset which is created in laboratory settings by human domain experts  with clean transcripts and real-world call data collected from a benefit service center that needs pre-processing and annotation. To annotate this large amount of unsupervised call data, a semi-supervised annotation method is proposed.

Working with the DS2 and Wav2Vec2 applications, we develop employee benefit-specific ASR systems, and compare their  speech transcription performance with AWS and Google commercial ASR systems. Finally, we evaluate the viability of using error prone ASR transcriptions as part of spoken language understanding (SLU) via one downstream  benefit-specific natural language understanding task - intent classification based on outputs generated from different ASR systems. 


This paper is organized as follows: Section~\ref{se:data} introduces the domain-specific dataset used to build ASR systems along with a semi-supervised annotation method. 
Section \ref{se:models} presents the methodologies for acoustic, language and punctuation models.  Section \ref{se:experiment} evaluates experimental results across different types of speech. Section \ref{se:discussion} concludes the paper and outlines future work.

\section{Training Data \label{se:data} }
In this paper, we target and use data collected from a health care benefit call center (named BSCD) which are focused on customers looking for help or support with company provided benefits such as health insurance. 





\subsection{Dataset \label{subse:data} }
To gain significant improvement from fine-tuning AMs, large volumes of high-quality annotated data are needed.  Unfortunately, manual transcription of large datasets is a time-consuming and expensive process, requiring trained human annotators and substantial amounts of supervision. In addition, this manual process cannot guarantee 100\% error-free transcription and the calls that are selected to be annotated may not contain benefit terms, and thus are not guaranteed to be useful for the domain-specific fine-tuning. 

Based on real users' chat-bot and call data, a set of representative utterances are created as  a `script' by experts. 13 people with a variety of backgrounds recorded data in quiet environments. However, this process is still time consuming and labor-intensive. In addition, the recorded artificial datasets lack  real-world variability. 

 
The BSCD call center is a rich resource of benefit-specific acoustic data which covers diverse topics, all demographic varieties and noise conditions. 
It is feasible to collect vast amounts of unsupervised acoustic datasets which lack  correct transcriptions. To label unsupervised datasets with little  human intervention, a semi-supervised annotation method is proposed in section \ref{subse:annot}.

\subsection{Data Pre-processing \label{subse:dataprep} }
In practice, most acoustic model training procedures expect that the training data comes in the form of relatively short utterances paired with associated transcripts, since long utterances have higher cost than short utterances in speech recognition\cite{pmlr-v48-amodei16}. 

All the recording audio files have duration less than 15 seconds.  However,  the call dataset ranges from several minutes to more than hours. The lengths of the call audio files make it impractical to train,  
so all the calls are split into shorter segments based on silence. Silence is defined as anything under -43 dBFS with duration longer than 800ms. 

All the audio files including call segments and recording data are presented in wav format, resampled to 16k Hz. We only keep audio files with duration in the range of 1.5 seconds  to 15 seconds, since short audios usually do not contain any useful information, while long audios have much higher cost during training. 
 

All the utterance scripts that are used to fine-tune the acoustic models should have the same character inventory as the pre-trained acoustic models which contains only alphabetic letters. The following filtering rules are applied to all the utterance scripts: 


\begin{itemize}
	\item All target text is converted to lowercase.
    \item Punctuation markers not pronounced in speech are eliminated (e.g. Words joined with hyphen '-' are marked as non hyphen lexical items).	
	\item Abbreviations are replaced with corresponding full-word forms based on the content (e.g. Carla Dr Athens becomes carla drive athens, Dr Pepper becomes doctor pepper). 

	\item Numeric strings and symbols that \textit{are} pronounced in speech are replaced with orthographic strings and words:
	\begin{itemize}
		\item Specific terms with numbers or symbols are replaced with specific orthographic strings (e.g. 401k becomes four o one k,  ad\&d becomes a d n d).  
		\item Numeric strings starting with punctuation marker \$ are dollar amounts (e.g. \$50 becomes fifty dollars, \$20.45 becomes twenty dollars forty five cents).
		\item Numeric strings ending with punctuation marker \% are numeric quantities (e.g.  50\% becomes fifty percent).
		\item Ordinal strings usually are dates (e.g. 21st becomes twenty first).
			\item Numbers with one, two or three digits are usually read as  cardinal numbers (e.g.  22 becomes twenty two; 156 becomes one hundred fifty six).
		\item   Four-digit numbers between 1930 and 2030 are considered as years (e.g. 2022 becomes two thousand twenty two).
		\item  Numbers with four or more digits (outside of the year range)  are usually street numbers, phone numbers, or social security numbers (e.g.  4680 becomes four six eight zero).
	\end{itemize}
	\end{itemize}



Mistakes can occur when converting scripts to alphabetic letters. For example, year 2022 can be said as 'two thousand twenty two'  or 'twenty twenty two'. Utterances with mistakes can be eliminated via the semi-supervised learning annotation method described in section \ref{subse:annot}. 






\subsection{Semi-supervised learning annotation \label{subse:annot} }

\begin{figure*}
	\centering
	\includegraphics[scale=0.38]{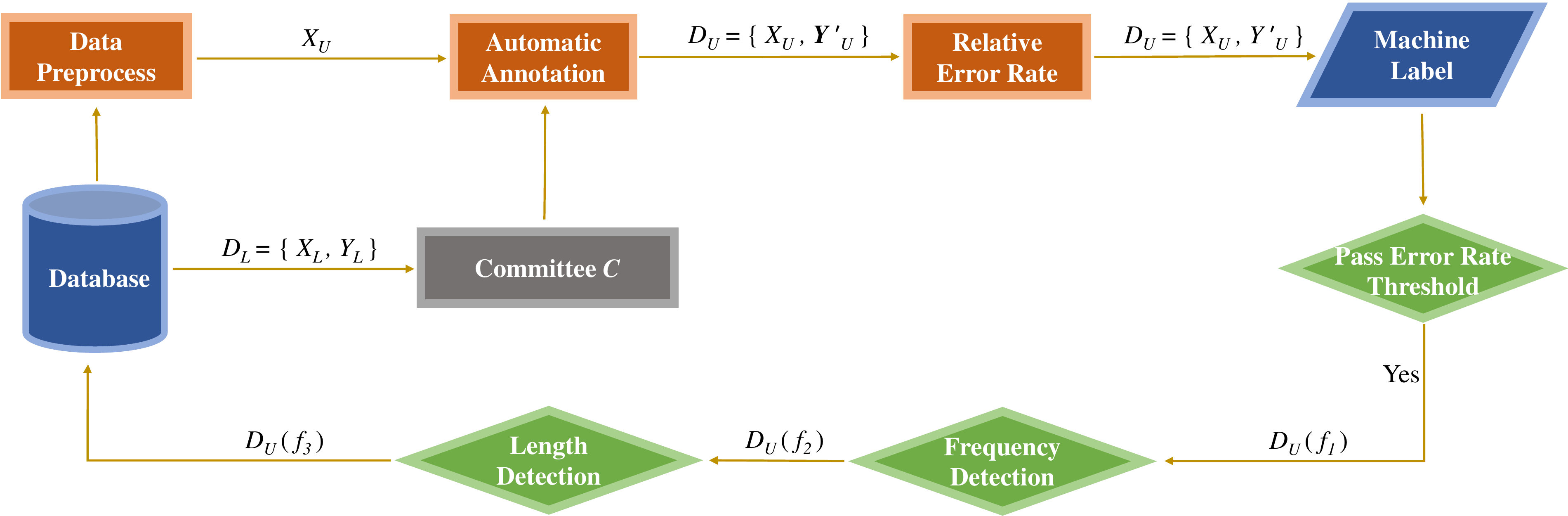}
	\caption{Semi-supervised learning annotation pipeline}
	\label{fig:dataprep}
\end{figure*}

Rather than relying on small, supervised training sets, a semi-supervised annotation method is proposed to bootstrap larger datasets, and reduce system development cost.  Semi-supervised annotation can be achieved with different components.  Figure \ref{fig:dataprep} illustrates the main components of a semi-supervised annotation process in an ASR system that can be implemented to bootstrap new data collection.

 The process takes as input a set of labeled examples $D_{L} = \{ X_L, Y_L\}$ and produces recognizer committee  $\boldsymbol{C}$ which is used to generate the rough transcriptions $Y^{\prime}_{U}$   
  for a larger set of unlabeled examples $X_{U}$. After a few filters, a relatively small set of newly labeled data $D_{U}(f)= \{X_{U}(f), Y^{\prime}_{U}(f) \} $ are adopted as part of  $D_{L}$ to retrain the system. The process is cyclical and iterative as every step is repeated to continuously improve the accuracy of the system and achieve a successful algorithm.

Recognizer committee $\boldsymbol{C}$ can include existing ASR systems or offline recognizers trained or fine-tuned using labeled audio data $D_{L}$. We employ DS2 and Wav2Vec2 models fine-tuned on $D_{L}$ along with AWS and Google commercial ASR systems to form recognizer  committee  $\boldsymbol{C}$ which automatically generates rough transcriptions $\boldsymbol{Y}_U  =  \{ Y_U^{\text{AWS} }, Y_U^{\text{Google}}, Y_U^{\text{DS2}} , Y_U^{\text{Wav2Vec2}}    \} $ for a massive amount of unannotated audio data $X_U$.

To cull salient utterances $X_U(f)$ with associated transcript $Y_U^{\prime}(f)$ from large unsupervised call data $X_U$,  we first need to find the most accurate transcript $Y_U^{\prime}$ from $\boldsymbol{Y}_U$.
In this step, a new metric `relative  error rate' is proposed.

For each unlabeled audio file $x_{u}(i)$  and associate transcript $y_{u}^{j}(i)$, the relative word error rate $wer^{jk}(i)$ and relative character error rate $cer^{jk}(i)$ are calculated by setting $y_{u}^{j}(i)$ as the target text and the transcript  $y_{u}^{k}(i)$ generated from  recognizer $k$  as the prediction transcript, where  $j$=$\{$AWS, Google, DS2, Wav2Vec2$\}$, and $k$ can be any one of the other three recognizers in the committee $\boldsymbol{C}$. 
The average $\bar{wer}^j(i) = \frac{1}{3} \sum_{k} wer^{jk}(i) $ and    $\bar{cer}^j(i) = \frac{1}{3} \sum_{k} cer^{jk}(i) $ are defined as  relative word error rate and relative character error rate  of  $y_{u}^{j}(i)$.

We normalize relative $\bar{wer}$ and  $\bar{cer}$ separately, 
then calculate the weighted arithmetic mean to get relative error rate $\bar{er}^j$ for all the four ASR systems separately. 
The `true' transcript for  $x_{u}(i)$ is set to be $y_{u}^{\prime}(i)$ with minimal relative error rate  $ \bar{er}^{\prime}(i) =min\{ \bar{er}^{\text{AWS} }(i), \bar{er}^{\text{Google} }(i), \bar{er}^{\text{DS2} }(i), \bar{er}^{\text{Wav2Vec2} }(i)  \} $. The label $Y_U^{\prime}$ are mixed transcripts produced from all the four different recognizers. 


A subset $D_U(f_1) = \{X_U(f_1), Y^{\prime}_U(f_1)\}$ is selected from $D_U = \{X_U, Y^{\prime}_U\}$ using an empirically determined threshold. An audio file  $x_{u}(i)$ with large relative error rate usually has loud background noise, overlap, heavy accent, or slurring which tends to lead to incorrect transcription ${y^{\prime}_u(i)}$.   Upper thresholds for  $\bar{wer}^{\prime}(i) $, $\bar{cer}^{\prime}(i)$ and  $\bar{er}^{\prime}(i) $ are empirically determined to make sure $x_{u}(i)$ has  accurate transcript. While $x_{u}(i)$ with relatively large $\bar{wer}^{\text{DS2}}(i) $, $\bar{cer}^{\text{DS2}}(i)$ and $\bar{wer}^{\text{Wav2Vec2}}(i) $, $\bar{cer}^{\text{Wav2Vec2}}(i)$ is useful to correct the wrong transcriptions generated by DS2 and Wav2Vec2 separately. So, lower thresholds for $\bar{wer}^{\text{DS2}}(i) $, $\bar{cer}^{\text{DS2}}(i)$ and $\bar{wer}^{\text{Wav2Vec2}}(i) $, $\bar{cer}^{\text{Wav2Vec2}}(i)$ are also determined. 
Utterances that fall into the determined thresholds are kept  in $D_U(f_1) = \{X_U(f_1), Y^{\prime}_U(f_1)\} $.  

The determined thresholds and the recognizers in   $\boldsymbol{C}$ may vary for each iteration. When the error rates of fine-tuned ASR outputs are low enough, we can remove the commercial ASR systems from  $\boldsymbol{C}$  or replace them with free ASR systems to reduce system development cost.


Selecting utterances by error rate threshold alone may lead to a disproportionate training dataset with frequent tokens over represented. To alleviate this problem, frequency detection can be implemented as one component of the pipeline. A threshold is determined to limit the number of  similar utterances which contain a set of the same key words in the datasets and make sure the filtered dataset $D_U(f_2)$  is diverse and has a larger vocabulary. 

Another constraint on data selection is the audio duration or character lengths. In this experiment, we only take audios with duration of 1.5 to 15 seconds as training data.  After transcription, utterances with character lengths less than 18 are also removed. For those eliminated utterances that are either too short or too long, reconstruction techniques can be used to reconstruct new audio segments to meet the length requirements.

The semi-supervised annotation approach directly relies on the transcripts generated by speech recognizers and the thresholds determined by experiments to cull sufficient quality data without any reliance on manual transcription \cite{semi}.

\section{Models \label{se:models}}

In this section, we explain the acoustic models, language models and punctuation models used in the ASR systems. 
\subsection{Acoustic Model \label{subse: acousitc} }

Compared to training a powerful AM from scratch, fine-tuning a pre-trained AM can be cheap in terms of data and training efforts. Since our experiment focuses on the employee benefit domain, we can improve the ASR performance through fine-tuning   pre-trained AMs on domain-specific datasets. Two popular open source speech-to-text engines: DS2 and Wav2Vec2 are employed. 

The core of DS2 is a recurrent neural network (RNN) trained with the Connectionist Temporal Classification (CTC) loss \cite{ctc}. It directly maps spectrograms to graphemes, and consists of 2 convolutional layers, 5 bidirectional RNN layers and a fully connected layer. 
DS2 is trained on 11,940 hours of labeled speech data containing 8 million utterances of English speech obtained from public sources and internal Baidu corpora. 
Noise superposition augmentation is employed to expand the training data even further.  Hundreds of hours of noise are added to 40\% of the utterances that are chosen at random to improve  robustness to noisy speech.

Wav2Vec2 is a fully convolutional model which is trained on large amounts of unlabeled audio data. It takes raw audio  as input and outputs latent speech representations which are input to the transformer.
After pre-training on unlabeled speech, the model can be fine-tuned on labeled data with a CTC loss to be used for downstream speech recognition tasks.  
Wav2Vec2 applies self-supervised pre-training to improve supervised speech recognition. It reduces the supervised data needed in real-world scenarios.

Three pre-trained Wav2Vec2 models are used in this experiment: BASE, LARGE, and LARGE-LV60. The
BASE and LARGE models have different transformer configurations setup. 
The BASE model contains 12 transformer blocks with 8 attention heads each, while the LARGE model contains 24 transformer blocks with 16 attention heads each. The BASE and LARGE pre-trained AMs are trained on  960 hours of unlabeled Librispeech audio (LS-960) \cite{7178964}. The LARGE-LV60 pre-trained AM uses audio data from LibriVox (LV-60k) and follows the pre-processing of \cite{libri} resulting in 53.2k hours of audio.



\subsection{Language Model \label{subse:language}}
Acoustic models can learn to produce readable character-level transcriptions; however, the acoustic model outputs tend to contain errors which occur on phonetically plausible renderings of English words including substitution of phonetically similar words, or misspellings due to irregularities in a language’s orthography.  Thus, integrating ASR systems with an external language model trained from  domain-specific text can further improve the ASR performance. 
A 5-gram KenLM  is trained on 10 million words of cleaned domain-specific documents  to decode the emissions from the acoustic model. A scorer is also trained as an external language model for DS2. The scorer is composed of a KenLM  and a trie data structure containing all words in the vocabulary.

\subsection{Punctuation Model \label{subse:punctuation}}

State-of-the-art speech recognition models
still produce raw word streams which  do not contain punctuation marks and capitalization.  Unformatted text is often difficult to read — not only for humans, but also for natural language processing tools. 
The presence of punctuation and capitalization can greatly improve the readability of automatic speech transcripts. 

A bidirectional recurrent neural network model with attention mechanism trained on 40M words of unsegmented text is used to restore 
missing inter-word punctuation marks: comma, period, question mark, exclamation mark, dash, colon, and semicolon \cite{tilk2016}. 

After inserting punctuation symbols and correctly capitalizing all words in ASR output, numbers and symbols addressed in section \ref{subse:dataprep} are also reversed to remove dysfluencies. 
We limit the problem scope to specific terms with numbers and symbols (e.g. `w two' becomes `w2', `covid nineteen' becomes `covid-19', `a d n d' becomes `ad\&d'). 

\section{Experiment \label{se:experiment}}

In this section, we empirically investigate the performance of the fine-tuned AMs along with external LMs on domain-specific audio data which contains high quality recorded artificial data and real telephone calls.   
To better assess the real-world applicability of our speech system, we also evaluate a natural language understanding task -  intent detection  - on error prone transcripts generated by different speech recognition systems.



\subsection{Experiment Setup \label{subse:setup}}
In this experiment, we used 57.6 hours of company internal call center data along with 1.8 hours of recording data  discussed in section \ref{subse:data}. The call center data is filtered down to  21.1 hours of call segments using the semi-supervised annotation method proposed in section \ref{subse:annot}. The training data statistics are summarized in Table 1. Compared to the recording data, the call segment data are easier to collect, and cover more unique words, diverse speakers, and more numbers of words/characters in each audio file. 


The test set contains  95 recording and 907 call segment audio files with approximate size 10 minutes and 75 minutes respectively. The call test data are randomly selected from company internal call segments, which covers  general utterances such as greetings and farewell. To better evaluate the speech system for benefit-specific audios, we  also created a subset of the call test data by selecting only those call segments  containing benefit specific terms, which yielded 565 audio files with  approximate size 48 minutes. 

All the test audios are manually  annotated by two annotators. They transcribe the same audio clip, that is typically about 6s (seconds) long with a range from 0.8s to 27.8s. A third annotator selects the better of the two transcriptions as ground truth to produce WER and CER. 

 
  \begin{table*}
 	\centering
 	\resizebox{.9\textwidth}{!}{
 		\begin{tabular}{l|cccccccc} 
 			\hline
 			\multirow{2}{*}{Data} &  Total & Training &  Total  & Unique  & Nb. Speakers  & Median &   Median   &  Median  \\  
 			&  Duration &  Duration   & Words & Words & or calls & Words & Characters & Duration \\ \hline
 			Recording &    1.8h & 1.8h & 13k  & 1.2k & 13 &  7  &   40 &   3.9s\\ 
 			Call &    57.6h &  21.1h &  246k & 5.8k & 864  &    11  &  56 & 3.7s \\    \hline
 	\end{tabular}}
 	\caption{Recording  and call segments training data statistics. }
 	\label{table:fusresults}
 \end{table*}

The four pre-trained acoustic models discussed in section \ref{subse: acousitc} are fine-tuned on the 23 hours of labeled speech data described in Table \ref{table:fusresults} with different numbers of epochs to reach a stable validation error rate and  different minibatch numbers to avoid out of memory issues.
The learning rate is chosen from  $ [ 1 \times 10^{-4}, 6 \times 10^{-4}  ] $ to yield fastest convergence. 
The external KenLM and scorer described in section \ref{subse:language} are trained for Wav2Vec2 and DS2 acoustic models separately on the same domain-specific text corpus.

\subsection{Results Evaluation \label{subse:result} }
The relative improvements gained from four pre-trained AMs and external LMs across three audio test sets are shown in table \ref{table: systemresults}. Speech recognition performance gains are observed by switching the  use of pre-trained model from DS2 to Wav2Vec2. These improvements are achieved without increasing fine-tuning data size. For Wav2Vec2 models, we observe that the use of larger pre-trained model leads to reductions in the WER and CER of the system.  For all the four fine-tuned acoustic models, an external LM improves the WER across all the test sets, with the largest relative WER improvement for the DS2 and Wav2Vec2-Base models and the smallest gain for Wav2Vec2-LARGE model.   
 
\begin{table}
	\centering
	\resizebox{0.75\textwidth}{!}{
		\begin{tabular}{l|cc|cc|cc} 
			\hline
			\multirow{2}{*}{System} &  \multicolumn{2}{c|}{Clean Recording} & \multicolumn{2}{c|}{Call} & \multicolumn{2}{c}{Call-Benefit}   \\  \cline{2-7} 
			&  WER & CER  & WER & CER & WER & CER   \\ \hline
			 DeepSpeech2  & 0.292  & 0.092 &  0.390    & 0.183 & 0.353 & 0.154 \\ \hline   
			Wav2Vec2-BASE  &     0.096 & 0.030    & 0.257  &0.129 & 0.205&  0.094\\ \hline
			Wav2Vec2-LARGE  &    \textbf{0.056} & \textbf{0.023}   & 0.194  &0.102 &0.152 & 0.075\\ \hline
			Wav2Vec2-LARG-LV60  & 0.069  & \textbf{0.023}  &  \textbf{0.189}  & \textbf{0.094}  &   \textbf{0.150}& \textbf{0.068}  \\ \hline
			DeepSpeech2 +LM 	& 0.063  & 0.028  &  0.262& 0.171  & 0.210 &  0.130\\ \hline 
			Wav2Vec2-BASE + LM & 0.057  & 0.024 & 0.165    & 0.098   &0.124 & 0.069 \\ \hline
			Wav2Vec2-LARGE + LM  & 0.043  & \textbf{0.021}& 0.154   & 0.093  &0.120 & 0.069 \\ \hline
			Wav2Vec2-LARG-LV60 + LM  &  \textbf{0.039}  & \textbf{0.021}  &  \textbf{0.138} & \textbf{0.083}  & \textbf{0.105} & \textbf{0.059} \\ \hline
	\end{tabular}}
	\caption{Comparison of WER/CER of  fine-tuned ASR systems across three  audio test sets. Best performance  bolded by column. }
	\label{table: systemresults}
\end{table}

 

\begin{table}
	\centering
	\resizebox{0.51\textwidth}{!}{
		\begin{tabular}{l|ccc } 
			\hline
			 System  &  Clean Recording  & Call  &  Call-Benefit  \\ \hline
			AWS API& 0.1446  &    \textbf{ 0.129}  &      0.113  \\ \hline
			 Google API & 0.075 &     0.260  &     0.219     \\ \hline
			Wav2Vec2    & \textbf{0.039}   & 0.138 &   \textbf{ 0.105} \\ \hline
	\end{tabular}}
	\caption{Comparison of WER  of three ASR systems across three audio test sets. Best performance bolded by column. Error rates are reported only for utterances with predictions given by all systems. }
	\label{table: systemresults3}
\end{table}


The transcripts carried out by  AWS and Google ASR systems are used as reference for  comparison. The commercial ASR outputs usually include symbols besides alphabetic letters for easy reading. To calculate WER, one human annotator follows the filtering rules in section \ref{subse:dataprep} and changes the commercial ASR outputs to  alphabetic letters only transcripts.   Three call segments with Google output `UnknownValueError' are removed from the test data. 
WER of the best fine-tuned ASR outputs (Wav2Vec2-LARG-LV60 + LM) along with two commercial ASR outputs across three audio test sets are reported in table \ref{table: systemresults3}.

For the general call test set, the AWS ASR system produces the lowest WER, but for the domain-specific audios including the recording test set and call-benefit test set, the best fine-tuned Wav2Vec2 model  outperforms the AWS ASR system, especially for recording data which contains many benefit specific terms.

Speech input industrial applications fulfill incoming user requests through the use of Spoken Language Understanding (SLU) \cite{7078634} which  applies NLU  tasks after ASR. The ASR system errors would propagate to the downstream NLU and degrade the performance \cite{jia-2020-deep}. 
In this experiment, we also focused on evaluating one of the NLU tasks - intent detection - using error prone ASR transcriptions. 

In general,  transcripts with lower WER produce  more accurate  intent prediction. 
 However, even if the ASR system does not correctly transcribe the input speech into text, the final intent result could be correct if the output of the recognition part preserves sufficient semantic information for intent prediction.

Table \ref{table:intent} presents the intent classification results based on three ASR outputs: AWS, Google, and Wav2Vec2 (Wav2Vec2-large-LV60 + LM).  Since Wav2Vec2 outputs are alphabetic letters only and the intent classification model is trained on regular text data, the punctuation model and reformatting described in section \ref{subse:punctuation} are applied to generate the Wav2Vec2-F (Formatting) transcripts, then feed them to the downstream intent classifier.

 The test utterance intents are labeled by two annotators based on the transcripts. Without knowing context, some of the utterance intents are not clear. Among 904 call utterances, 626 have meaningful intent. Among 95 recording utterances, 83 have meaningful intent. Table \ref{table:intent} shows the WER for the test data and the number of correct intent predictions based on different speech recognition system outputs. The intent predictions based on the manual transcripts are also calculated as reference. The intent classifier model works good for recording transcripts, but for call segment transcripts the accuracy is around 77.6\%.

\begin{table*} 
	\centering
	\resizebox{0.72\textwidth}{!}{
		\begin{tabular}{lc|ccccc} 
			\hline
			Type & Error & \textit{Transcripts} & AWS & Google & Wav2Vec2 & Wav2Vec2-F   \\ \hline
			\multirow{2}{*}{Recording} &  WER  &\textit{0.000} & 0.136 & 0.070  &  \textbf{0.037} &  \textbf{0.037} \\ \cline{2-7} 
			& Intent  & \textit{83} & 75 & 78  &  \textbf{81}  &  \textbf{81}   \\ \hline
			\multirow{2}{*}{Call} &   WER & \textit{0.000}  &  \textbf{0.109} &0.243  & 0.119 & 0.119 \\ \cline{2-7} 
			&Intent  & \textit{486} & 481 & 388 & 481 &  \textbf{488} \\ \hline
	\end{tabular}}
	\caption{WER and the number of correct intent predictions based on different speech recognition system outputs. Best performance bolded by column.}
	\label{table:intent}
\end{table*}

For the recording test set, the intent prediction based on fine-tuned Wav2Vec2 outputs performs best. For the call segment test set,  AWS ASR output has the lowest WER but its intent classification performance is the same as  Wav2Vec2. The number of correct intent predictions based on  Wav2Vec2-F is even higher than AWS and  Wav2Vec2 direct alphabetic letters only transcript.  In addition, the intent classification results between fine-tuned ASR and human transcriptions are very similar. 
Therefore, although the intent classifier model can suffer from  erroneous speech systems output,  the domain-specific fine-tuned ASR transcriptions often contain enough accurate terms which are crucial to the meaning of the utterance that the intent classifier can still produce the  correct intents.




\section{Discussion \label{se:discussion}}


This paper presents a thorough analysis of how to build an end-to-end ASR system using company internal audio data. The insights we gleaned from this investigation provide hints on how companies could potentially adapt such pre-trained acoustic models and semi-supervised annotation to build their own end-to-end ASR system which is able to outperform the commercial ASR systems for domain-specific speech. 

However, building an ASR system through fine-tuning  pre-trained acoustic models still requires large enough supervised data to achieve reliable performance, and acoustic model performance is sensitive to the accuracy of training transcriptions. At the beginning of the process, humans are needed to get high quality training corpus.  
In this experiment, recording data is used to initiate the process. It  takes time  and effort at the beginning, but the acoustic models can be easily refreshed  and improved as the training corpus size and the ASR system transcript quality increased with more iterations.  For  domain-specific SLU tasks, the use of domain-specific fine-tuned ASR output can outperform the use of commercial ASR transcriptions even when the fine-tuned model has higher WER, and it can even reach a similar performance with the use of  human transcriptions.

Building their own ASR system   can better protect personal identifiable information such as social security numbers and birth date which are commonly occurred in call center data. In addition, the cost spent on ASR system can be significantly reduced compared to employing commercial ASR systems especially for call centers with high call volumes daily.

\section{Conclusion \label{se:conclusion}}
In this paper, we introduce a fully automated procedure to select and annotate unsupervised audio data. This process allows us to easily improve transcript quality and increase  training corpus size,  therefore updating the ASR systems  with as little human intervention and as often as needed.

Four pre-trained acoustic models are fine-tuned on internal company corpora, and KenLM  is incorporated to further improve performance on domain-specific speech. The fine-tuned Wav2Vec2-LARG-LV60 + LM has lower WER for domain-specific speech and more accurate intent predictions compared to AWS and Google commercial speech recognition systems.  

For future work, we would like to explore and evaluate ASR system on more languages such as Chinese and Spanish, investigate ASR error robust downstream NLP systems, and compare with an end-to-end SLU system performance.  

\bibliography{Speech2Text}		
\bibliographystyle{ieeetr}
 
%
%


\end{document}